\documentclass[11pt,a4paper]{article}
\usepackage[hyperref]{emnlp-ijcnlp-2019}
\usepackage{times}
\usepackage{latexsym}
\usepackage{graphicx}
\usepackage{amsmath}
\usepackage{amssymb}
\usepackage{url}
\usepackage{algorithm2e}
\usepackage{color}
\usepackage{multirow}

\aclfinalcopy % Uncomment this line for the final submission

\renewcommand{\vec}[1]{\mathbf{#1}}

\title{Word class flexibility: A deep contextualized approach}

\author{Bai Li$^{1,4}$, Guillaume Thomas$^{2}$, Yang Xu$^{1,3,4}$, Frank Rudzicz$^{1,4,5}$ \\
$^1$ Department of Computer Science, University of Toronto, Toronto, Canada \\
$^2$ Department of Linguistics, University of Toronto, Toronto, Canada \\
$^3$ Cognitive Science Program, University of Toronto, Toronto, Canada \\
$^4$ Vector Institute for Artificial Intelligence, Toronto, Canada \\ 
$^5$ St. Michael's Hospital, Toronto, Canada \\
\texttt{\{bai, yangxu, frank\}@cs.toronto.edu} \\ 
\texttt{guillaume.thomas@utoronto.ca}
}

\date{}

\newcommand \NumLanguages{37 }

\begin{document}
\maketitle
\begin{abstract}
Word class flexibility refers to the phenomenon whereby a single word form is used across different grammatical categories. Extensive work in linguistic typology has sought to characterize word class flexibility across languages, but quantifying this phenomenon accurately and at scale has been fraught with difficulties. We propose a principled methodology to explore regularity in word class flexibility. Our method builds on recent work in contextualized word embeddings to quantify semantic shift between word classes (e.g., noun-to-verb, verb-to-noun), and we apply this method to \NumLanguages languages\footnote{Code and data to reproduce the experimental findings are available at: \url{https://github.com/SPOClab-ca/word-class-flexibility}.}. We find that contextualized embeddings not only capture human judgment of  class variation within words in English, but also uncover shared tendencies in class flexibility across languages. Specifically, we find greater semantic variation when flexible lemmas are used in their dominant word class, supporting the view that word class flexibility is a directional process. Our work highlights the utility of deep contextualized models in linguistic typology.
\end{abstract}

\section{Introduction}

In natural languages, lexical items can often be used in multiple word classes without overt changes in word form. For instance, the word \textit{buru} in Mundari can be used as a noun to denote `mountain', or as a verb to denote `to heap up' \citep{evans-osada}. Known as word class flexibility, this phenomenon is considered one of the most challenging topics in linguistic typology \citep{EvansLevinson2009}. We present a computational methodology to quantify the regularity in word class flexibility across languages. 

There is an extensive literature on how languages vary in word class flexibility, either directly \citep{Hengeveld1992, VogelComrie2000, vanLierRijkhoff2013} or through related notions such as word class conversion (with zero-derivation) \citep{Vonen1994, Don2003, BauerValera2005a, Manova2011, StekauerEtAl2012}. However, existing studies tend to rely on analyses of small sets of lexical items that may not be representative of word class flexibility in the broad lexicon. Critically lacking are systematic analyses of word class flexibility across many languages, and existing typological studies have only focused on qualitative comparisons of word class systems. 

We take to our knowledge the first step towards computational quantification of word class flexibility in \NumLanguages languages, taken from the Universal Dependencies project \citep{Zeman2019}. We focus on lexical items that can be used both as nouns and as verbs, i.e., noun-verb flexibility. This choice is motivated by the fact that the distinction between nouns and verbs is the most stable in word class systems across languages: if a language makes any distinction between word classes at all, it will likely be a distinction between nouns and verbs \citep{Hengeveld1992, Evans2000, Croft2003}. However, our understanding of cross-linguistic regularity in noun-verb flexibility is impoverished.

We operationalize word class flexibility as a property of lemmas. We define a lemma as flexible if some of its occurrences are tagged as nouns and others as verbs. Flexible lemmas are sorted into noun dominant lemmas, which occur more frequently as nouns, and verb dominant lemmas that occur more frequently as verbs. Our methodology builds on contextualized word embedding models (e.g., ELMo \citep{elmo} and BERT \citep{bert}) to quantify semantic shift between grammatical classes of a lemma, within a single language. This methodology can also help quantify metrics of flexibility in the lexicon across  languages.

We use our methodology to address one of the most fundamental questions in the study of word class flexibility: should this phenomenon be analyzed as a directional word-formation process similar to derivation, or as a form of underspecification? Derived words are commonly argued to have a lower frequency of use and a narrower range in meaning compared to their base \citep{Marchand1964, Iacobini2000}. If word class flexibility is a directional process, we should expect that flexible lemmas are subject to more semantic variation in their dominant word class than in their less frequent class. We also test the claim that noun-to-verb flexibility  involves  more  semantic shift  than  verb-to-noun flexibility.  While previous work has explored these questions, it remains challenging to quantify semantic shift and semantic variation, particularly across different languages.

We present a novel probing task that reveals the ability of deep contextualized models to capture semantic information across word classes. Our utilization of deep contextual models predicts human judgment on the spectrum of noun-verb flexible usages including homonymy (unrelated senses), polysemy (different but related senses), and word class flexibility. We find that BERT outperforms ELMo and non-contextual word embeddings, and that the upper layers of BERT capture the most semantic information, which resonates with existing probing studies \citep{tenney-probing}.

\section{Related work and assumptions}

\subsection{Types of flexibility}

The phenomenon of word class flexibility has been analyzed in different ways. One way is to assume the existence of underspecified word classes. For instance, \citet{Hengeveld2013} claims that basic lexical items in Mundari belong to a single class of \textit{contentives} that can be used to perform all the functions associated with nouns, verbs, adjectives or adverbs in a language like English. Alternatively, word class flexibility can be analyzed as a form of conversion, i.e., as a relation between words that have the same form and closely related senses but different word classes, such as \textit{a fish} and \textit{to fish} in English \citep{Adams1973}. Conversion has been analyzed as a derivational process that relates different lexemes \citep{Jespersen1924, Marchand1969, QuirkEtAl1985}, or as a property of lexemes whose word class is underspecified \citep{Farell2001, BarnerBale2002}. We use word class flexibility as a general term that subsumes these different notions. This allows us to assess whether there is evidence that word class flexibility should be characterized as a directional word formation process, rather than as a form of underspecification.

\subsection{Homonymy and polysemy}

\label{sec:polysemy-homonymy}

Word class flexibility has often been analyzed in terms of homonomy and polysemy \citep{ValeraRuz2020}. Homonymy is a relation between lexemes that share the same word form but are not semantically related \citep[][p.80]{Cruse1986}. Homonyms may differ in word class, such as \textit{ring} `a small circular band' and \textit{ring} `make a clear resonant or vibrating sound.' Polysemy is defined as a relation between different senses of a single lexeme (\textit{ibid.}). Insofar as the nominal and verbal uses of flexible lexical items are semantically related, one may argue that word class flexibility is similar to polysemy, and must be distinguished from homonymy. In practice,  homonymy and polysemy exist on a continuum, so it is difficult to apply a consistent criterion to differentiate them \citep{tuggy-polysemy}. As a consequence, we will not attempt to tease homonymy apart from word class flexibility.

Regarding morphology, word class flexibility excludes pairs of lexical items that are related by overt derivational affixes, such as \textit{to act}/\textit{an actor}. In such cases, word class alternations can be attributed to the presence of a derivational affix, and are therefore part of regular morphology. In contrast, we allow tokens of flexible lexical items to differ in inflectional morphology.

\subsection{Directionality of class conversion}

\label{sec:related-work-directionality}

Word class flexibility can be analyzed either as a static relation between nominal and verbal uses of a single lexeme, or as a word formation process related to derivation. The merits of each analysis have been extensively debated in the literature on conversion \citep[see e.g.,][]{Farell2001, Don2005}. One of the objectives of our study is to show that deep contextualized language models can be used to help resolve this debate. A hallmark of derivational processes is their directionality. Direction of derivation can be established using several synchronic criteria, among which are the principles that a derived form tends to have a lower frequency of use and a smaller range of senses than its base \citep{Marchand1964, Iacobini2000}. In languages where word class flexibility is a derivational process, one should therefore expect greater semantic variation when flexible lemmas are used in their dominant word class---an important issue that we verify  with our methodology.

Semantic variation has been operationalized in several ways. \citet{kisselew} uses an entropy-based metric, while \citet{balteiro} and \citet{bram-thesis} measure semantic variation by counting the number of different noun and verb senses in a dictionary. The latter study found that the more frequent word class has greater semantic variation at a rate above random chance. Here we propose a novel metric based on contextual word embeddings to compare the amount of semantic variation of flexible lemmas in their dominant and non-dominant grammatical classes. Differing from existing methods, our metric is validated explicitly on human judgements of semantic similarity, and can be applied to many languages without the need for dictionary resources.

\subsection{Asymmetry in semantic shift}

\label{sec:related-work-semantic-shift}

If word class flexibility is a directional process, a natural question is whether derived verbs stand in the same semantic relation to their base as derived nouns. The literature on conversion suggests that there might be significant differences between these two directions of derivation. In English, verbs that are derived from nouns by conversion have been argued to describe events that include the noun's denotation as a participant (e.g. \textit{hammer}, `to hit something with a hammer') or as a spatio-temporal circumstance (\textit{winter} `to spend the winter somewhere'). \citet{clark-clark} argue that the semantic relations between denominal verbs and their base are so varied that they cannot be given a unified description. In comparison, when the base of conversion is a verb, the derived noun most frequently denotes an event of the sort described by the verb (e.g. \textit{throw} `the act of throwing something'), or the result of such an act (e.g. \textit{release} `state of being set free') \citep{Jespersen1942, Marchand1969, Cetnarowska1993}. This has led some authors to suggest that verb to noun conversion in English involves less semantic shift than noun to verb conversion \citep[][p.22]{Bauer2005}. Here we consider a new metric of semantic shift based on contextual embeddings, and we use this metric to test the hypothesis that the expected semantic shift involved in word class flexibility is greater for noun dominant lexical items (as compared to verb dominant lexical items) in our sample of languages. As we will show, this proposal is consistent with the empirical observation that verb-to-noun conversion is statistically more salient than noun-to-verb conversion.

\subsection{Contextualized language models}

Deep contextualized language models take a sequence of natural language tokens and produce a sequence of context-sensitive embeddings for each token. These embeddings can be used in a variety of downstream tasks and have achieved state-of-the-art performance in many of them. There are many models that generate contextual embeddings, generally trained with unsupervised learning using a large corpus. In particular, ELMo \citep{elmo} uses a left-to-right and a right-to-left LSTM trained to minimize perplexity across a large corpus. To generate contextual embeddings, it feeds the sentence through both LSTMs and concatenates the left-to-right and right-to-left LSTM states. BERT \citep{bert} uses 12 layers of the Transformer module \citep{transformer} and is pre-trained on a large corpus using two tasks: masked language modeling to predict randomly masked tokens from context, and next sentence prediction to predict whether two sentences are contiguous in the original text.

Both ELMo and BERT can be adapted to non-English languages without modification. The authors of BERT trained multilingual BERT (mBERT) by concatenating Wikipedia for 104 languages. There are models designed specifically for multilingual situations: XLM \citep{xlm} and XLM-R \citep{xlmr} are similar to BERT, but include an additional pre-training objective that leverages parallel text.

Typically, BERT is used in combination with task-specific modules and the parameters fine-tuned using domain data. Here we use contextual embeddings without fine-tuning. Probing experiments revealed that BERT embeddings contain semantic information beyond static embeddings, especially in the upper layers \citep{tenney-probing}, and this information is demonstrably useful for word sense disambiguation \citep{wsd-bert1, wsd-bert2}.

\section{Identification of word class flexibility}

\subsection{Definitions}
\label{sec:definitions}

A lemma is {\em flexible} if it can be used both as a noun and as a verb. To reduce noise, we require each lemma to appear at least 10 times and at least 5\% of the time as the minority class to be considered flexible. The {\em inflectional paradigm} of a lemma is the set of words that have the lemma.

A flexible lemma is {\em noun (verb) dominant} if it occurs more often as a noun (verb) than as a verb (noun). This is merely an empirical property of a lemma: we do not claim that the base POS should be determined by frequency. The {\em noun (verb) flexibility} of a language is the proportion of noun (verb) dominant lemmas that are flexible.

\subsection{Datasets and preprocessing}

Our experiments require corpora containing part-of-speech annotations. For English, we use the British National Corpus (BNC), consisting of 100M words of written and spoken English from a variety of sources \citep{bnc}. Root lemmas and POS tags are provided, and were generated automatically using the CLAWS4 tagger \citep{claws4}. For our experiments, we use BNC-baby, a subset of BNC containing 4M words.

For other languages, we use the Universal Dependencies (UD) treebanks of over 70 languages, annotated with lemmatizations, POS tags, and dependency information \citep{Zeman2019}. We concatenate the treebanks for each language and use the languages that have at least 100k tokens.

The UD treebanks are too small for our contextualized experiments and are not matched for content and style, so we supplement them with Wikipedia text\footnote{We use Wikiextractor to extract text from Wikimedia dumps: \url{https://github.com/attardi/wikiextractor}.}. For each language, we randomly sample 10M tokens from Wikipedia; we then use UDPipe 1.2 \citep{udpipe} to tokenize the text and generate POS tags for every token. We do not use the lemmas provided by UDPipe, but instead use the lemma merging algorithm to group lemmas.

\subsection{Lemma merging algorithm}

\begin{table}[t]
    \small
    \centering
\begin{tabular}{|l|l|l|l|l|l|}
\hline
\textbf{Language} & \textbf{Nouns} & \textbf{Verbs} & \textbf{\begin{tabular}[c]{@{}l@{}}Noun\\ flexibility\end{tabular}} & \textbf{\begin{tabular}[c]{@{}l@{}}Verb\\ flexibility\end{tabular}} \\ \hline
Arabic & 1517 & 299 & 0.076 & 0.221 \\ \hline
Bulgarian & 786 & 343 & 0.039 & 0.047 \\ \hline
Catalan & 1680 & 590 & 0.039 & 0.147 \\ \hline
Chinese & 1325 & 634 & 0.125 & 0.391 \\ \hline
Croatian & 1031 & 370 & 0.042 & 0.062 \\ \hline
Danish & 324 & 216 & 0.108 & 0.269 \\ \hline
Dutch & 958 & 441 & 0.077 & 0.188 \\ \hline
English & 1700 & 600 & 0.248 & 0.472 \\ \hline
Estonian & 1949 & 592 & 0.032 & 0.115 \\ \hline
Finnish & 1523 & 631 & 0.028 & 0.136 \\ \hline
French & 1844 & 649 & 0.062 & 0.257 \\ \hline
Galician & 802 & 334 & 0.031 & 0.135 \\ \hline
German & 4239 & 1706 & 0.049 & 0.229 \\ \hline
Hebrew & 850 & 315 & 0.111 & 0.321 \\ \hline
Indonesian & 572 & 243 & 0.052 & 0.128 \\ \hline
Italian & 2227 & 770 & 0.067 & 0.256 \\ \hline
Japanese & 1105 & 417 & 0.178 & 0.566 \\ \hline
Korean & 1890 & 1003 & 0.026 & 0.048 \\ \hline
Latin & 1090 & 885 & 0.056 & 0.122 \\ \hline
Norwegian & 1951 & 636 & 0.072 & 0.259 \\ \hline
Old Russian & 527 & 416 & 0.034 & 0.060 \\ \hline
Polish & 2054 & 1084 & 0.069 & 0.427 \\ \hline
Portuguese & 1711 & 638 & 0.037 & 0.185 \\ \hline
Romanian & 1809 & 740 & 0.060 & 0.151 \\ \hline
Slovenian & 746 & 316 & 0.068 & 0.123 \\ \hline
Spanish & 2637 & 873 & 0.046 & 0.202 \\ \hline
Swedish & 784 & 384 & 0.038 & 0.109 \\ \hline \hline
\multicolumn{5}{|l|}{{\bf Excluded Languages}} \\ \hline
Ancient Greek & 1098 & 1022 & 0.015 & 0.026 \\ \hline
Basque & 650 & 247 & 0.020 & 0.105 \\ \hline
Czech & 5468 & 2063 & 0.004 & 0.011 \\ \hline
Hindi & 1364 & 133 & 0.019 & 0.135 \\ \hline
Latvian & 1159 & 603 & 0.022 & 0.061 \\ \hline
Persian & 1125 & 47 & 0.010 & 0.234 \\ \hline
Russian & 3909 & 1760 & 0.005 & 0.024 \\ \hline
Slovak & 488 & 281 & 0.006 & 0.011 \\ \hline
Ukrainian & 659 & 238 & 0.006 & 0.029 \\ \hline
Urdu & 722 & 51 & 0.018 & 0.216 \\ \hline
\end{tabular}
    \caption{Noun and verb flexibility for 37 languages with at least 100k tokens in the UD corpus. We  include the 27 languages with over 2.5\% noun and verb flexibility; 10 languages are excluded from further analysis.}
    \label{tab:ud-frequency}
\end{table}

The UD corpus provides lemma annotations for each word, but these lemmas are insufficient for our purposes because they do not always capture instances of flexibility. In some languages, nouns and verbs are lemmatized to different forms by convention. For example, in French, the word \textit{voyage} can be used as a verb (\textit{il voyage} `he travels') or as a noun (\textit{un voyage} `a trip'). However, verbs are lemmatized to the infinitive \textit{voyager}, whereas nouns are lemmatized to the singular form \textit{voyage}. Since the noun and verb lemmas are different, it is not easy to identify them as having the same stem.

The different lemmatization conventions of French and English reflect a more substantial linguistic difference. French has a stem-based morphology, in which stems tend to occur with an inflectional ending. By contrast, English has a word-based morphology, where stems are commonly used as free forms \citep{Kastovsky2006}. This difference is relevant to the definition of word class flexibility: in stem-based systems, flexible items are stems that may not be attested as free forms \citep[][p.14]{BauerValera2005b}.

We propose a heuristic algorithm to capture stem-based flexibility as well as word-based flexibility. The key observation is that the inflectional paradigms of the noun and verb forms often have some words in common (such is the case for \textit{voyager}). Thus, we merge any two lemmas whose inflectional paradigms have a nonempty intersection. This is implemented with a single pass through the corpus, using the union-find data structure: for every word, we call UNION on the inflected form and the lemmatized form.

Using this heuristic, we can identify cases of flexibility that do not share the same lemma in the UD corpus (Table \ref{tab:ud-frequency}). This method is not perfect, and is unable to identify cases of stem-based flexibility where the inflectional paradigms don't intersect, for example in French, \textit{chant} `song' and \textit{chants} `songs' are not valid inflections of the verb \textit{chanter} `to sing'. There are also false positives that cause two unrelated lemmas to be merged if their inflectional paradigms intersect, for example, \textit{avions} (plural form of \textit{avion} `airplane') happens to have the same form as \textit{avions} (first person plural imperfect form of \textit{avoir} `to have').

\section{Methodology and evaluation}

\subsection{Probing test of contextualized model}

Deep contextual embeddings can capture a variety of information other than semantics, which can introduce noise into our results, for example: the lexicographic form of a word, syntactic position, etc. In order to compare different contextual language models on how well they capture semantic information, we perform a probing test of how accurate the models can capture human judgements of  word sense similarity.

We begin with a list of the 138 most frequent flexible words in the BNC corpus. Some of these words are flexible (e.g., \textit{work}), while others are homonyms (e.g., \textit{bear}). For each lemma, we get five human annotators from Mechanical Turk to make a sentence using the word as a noun, then make a sentence using the word as a verb, then rate the similarity of the noun and verb senses on a scale from 0 to 2. The sentences are used for quality assurance, so that ratings are removed if the sentences are nonsensical. We will call the average human rating for each word the {\em human similarity score}.

Next, we evaluate each layer of ELMo, BERT, mBERT, and XLM-R\footnote{We use the models `bert-base-uncased', `bert-base-multilingual-cased', and `xlm-roberta-base' from \citet{huggingface}.} on correlation with the human similarity score. That is, we compute the mean of the contextual vectors for all noun instances of the given word in the BNC corpus, the mean across all verb instances, then compute the cosine distance between the two mean vectors as the model's similarity score. Finally, we evaluate the Spearman correlation of the human and model's similarity scores for 138 words: this score measures the model's ability to gauge the level of semantic similarity between noun and verb senses, compared to human judgements.

For a baseline, we do the same procedure using non-contextual GloVe embeddings \citep{glove}. Note that while all instances of the same word have a static embedding, different words that share the same lemma still have different embeddings (e.g., \textit{work} and \textit{works}), so that the baseline is not trivial.

\begin{figure}
    \centering
    \includegraphics[width=\linewidth]{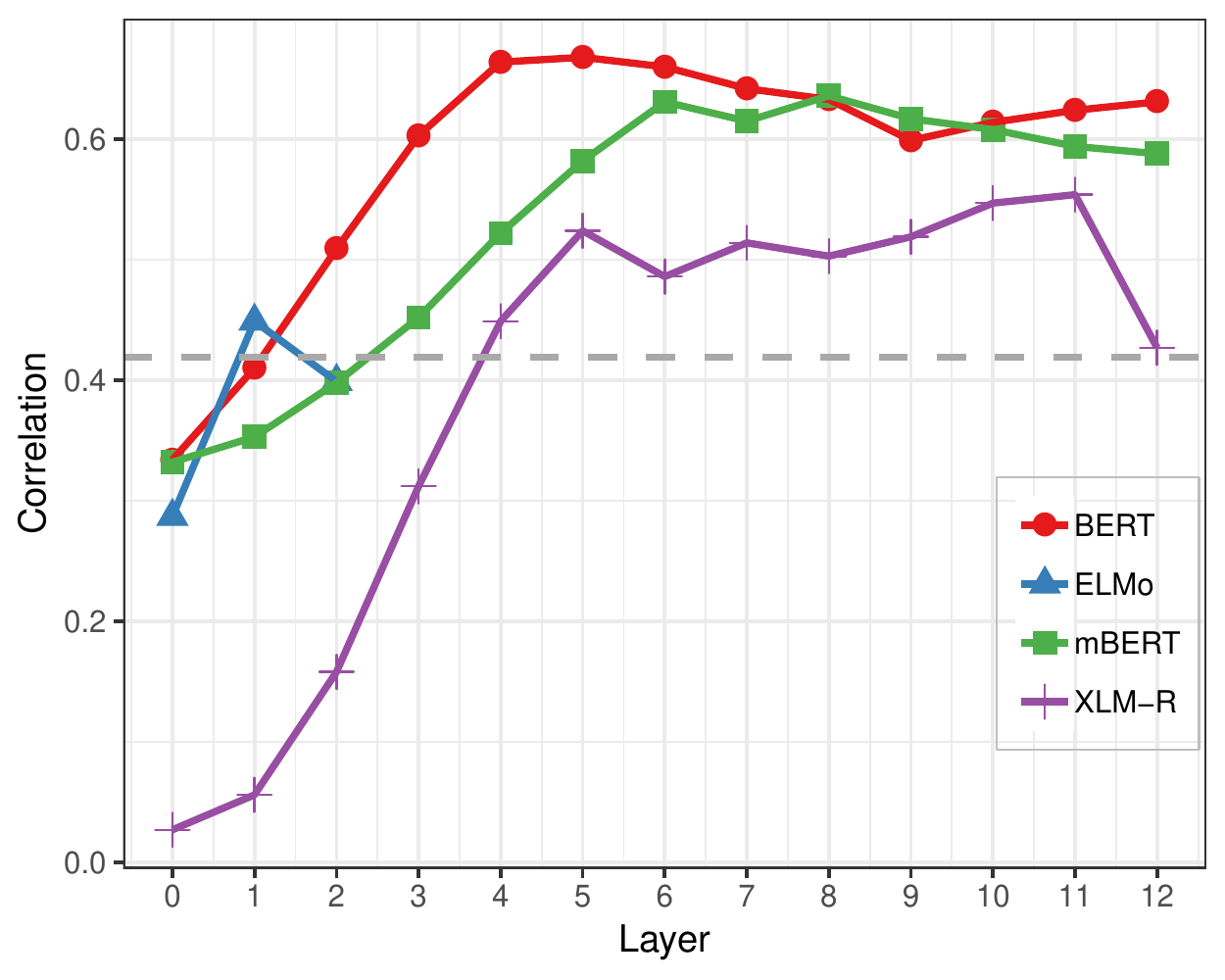}
    \caption{Spearman correlations between human and model similarity scores for ELMo, BERT, mBERT, and XLM-R. The dashed line is the baseline using static GloVe embeddings.}
    \label{fig:mturk-correlations}
\end{figure}

The correlations are shown in Figure \ref{fig:mturk-correlations}. BERT and mBERT are better than ELMo and XLM-R at capturing semantic information, in all transformer models, the correlation increases for each layer up until layer 4 or so, and after this point, adding more layers neither improves nor degrades the performance. Thus, unless otherwise noted, we use the final layers of each model for downstream tasks.

\begin{figure}[t]
    \centering
    \includegraphics[width=\linewidth]{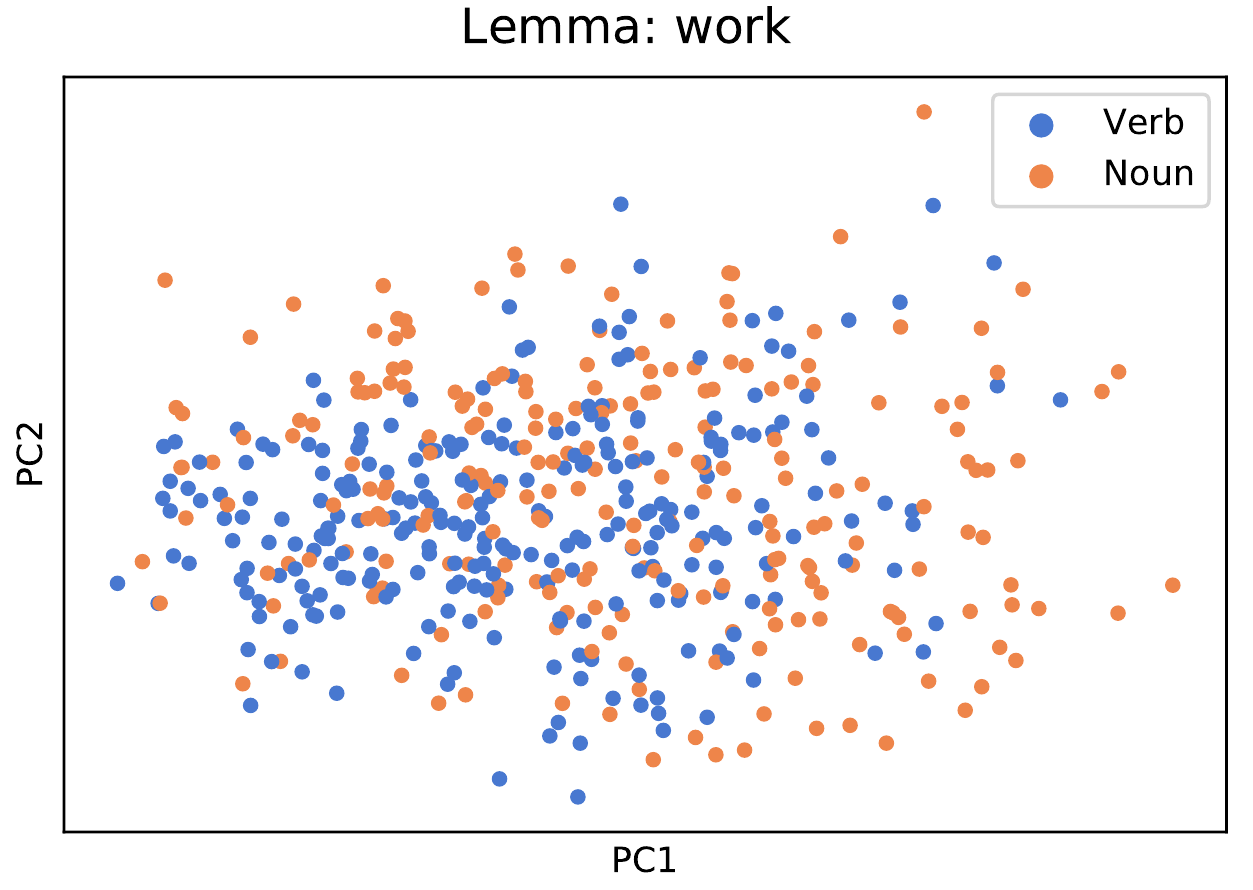}
    \includegraphics[width=\linewidth]{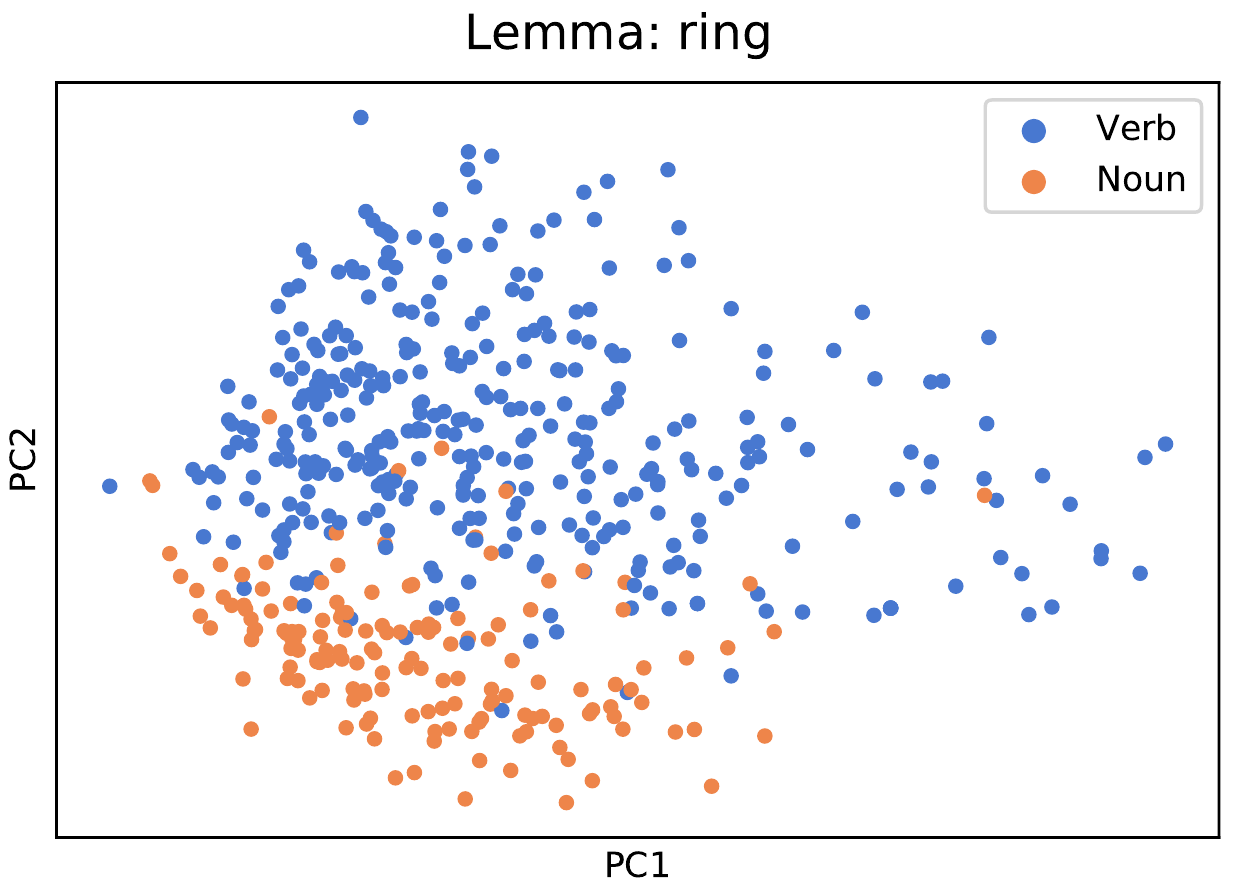}
    \caption{PCA plot of BERT embeddings for the lemmas ``work'' (high similarity between noun and verb senses) and ``ring'' (low similarity).}
    \label{fig:bert-work}
\end{figure}

Figure \ref{fig:bert-work} illustrates the contextual distributions for two lemmas on the opposite ends of the noun-verb similarity spectrum: \textit{work} (human similarity score: 2) and \textit{ring} (human similarity score: 0). We apply PCA to the BERT embeddings of all instances of each lemma in the BNC corpus. For \textit{work}, the noun and verb senses are very similar and the distributions have high overlap. In contrast, for \textit{ring}, the most common noun sense (`a circular object') is etymologically and semantically unrelated to the most common verb sense (`to produce a resonant sound'), and accordingly, their distributions have very little overlap.

\begin{table*}[t]
    \centering
    \begin{tabular}{|l||l|l||l|l||l|l|}
    \hline
    \textbf{Language} & \textbf{N$\to$V shift} & \textbf{V$\to$N shift} & \textbf{\begin{tabular}[c]{@{}l@{}}Noun\\ variation\end{tabular}} & \textbf{\begin{tabular}[c]{@{}l@{}}Verb\\ variation\end{tabular}} & \textbf{\begin{tabular}[c]{@{}l@{}}Majority\\ variation\end{tabular}} & \textbf{\begin{tabular}[c]{@{}l@{}}Minority\\ variation\end{tabular}} \\ \hline
    Arabic & 0.098 & 0.109 & 8.268 & 8.672$^{***}$ & 8.762$^{***}$ & 8.178 \\ \hline
    Bulgarian & 0.146 & 0.136 & 8.267 & 8.409 & 8.334 & 8.341 \\ \hline
    Catalan & 0.165 & 0.169 & 8.165 & 8.799$^{***}$ & 8.720$^{***}$ & 8.244 \\ \hline
    Chinese & 0.072 & 0.070 & 7.024 & 7.212$^{***}$ & 7.170$^{***}$ & 7.067 \\ \hline
    Croatian & 0.093 & 0.144$^{**}$ & 8.149 & 8.109 & 8.219$^{**}$ & 8.037 \\ \hline
    Danish & 0.103 & 0.110 & 8.245 & 8.338 & 8.438$^{***}$ & 8.146 \\ \hline
    Dutch & 0.146 & 0.174 & 7.716 & 8.786$^{***}$ & 8.354$^{*}$ & 8.148 \\ \hline
    English & 0.175$^{*}$ & 0.160 & 8.035 & 8.624$^{***}$ & 8.390$^{***}$ & 8.268 \\ \hline
    Estonian & 0.105 & 0.103 & 7.800 & 7.902 & 8.022$^{**}$ & 7.679 \\ \hline
    Finnish & 0.100 & 0.114 & 7.972 & 7.854 & 8.181$^{***}$ & 7.644 \\ \hline
    French & 0.212 & 0.204 & 8.189 & 9.472$^{***}$ & 9.082$^{***}$ & 8.578 \\ \hline
    Galician & 0.111 & 0.117 & 7.922 & 8.340$^{***}$ & 8.137 & 8.127 \\ \hline
    German & 0.382 & 0.355 & 8.078 & 9.758$^{***}$ & 9.096$^{**}$ & 8.740 \\ \hline
    Hebrew & 0.121 & 0.130 & 8.096 & 9.116$^{***}$ & 8.574 & 8.638 \\ \hline
    Indonesian & 0.034 & 0.048 & 7.100 & 7.076 & 7.076 & 7.101 \\ \hline
    Italian & 0.207 & 0.184 & 8.520 & 9.345$^{***}$ & 9.149$^{***}$ & 8.716 \\ \hline
    Japanese & 0.061 & 0.057 & 7.419$^{***}$ & 7.173 & 7.309 & 7.283 \\ \hline
    %Korean & 0.087 & 0.079 & 7.685$^{***}$ & 7.285 & 7.592$^{*}$ & 7.377 \\ \hline
    Latin & 0.092 & 0.139$^{***}$ & 7.920$^{***}$ & 7.710 & 7.905$^{***}$ & 7.724 \\ \hline
    Norwegian & 0.133 & 0.132 & 8.112 & 8.336$^{***}$ & 8.332$^{***}$ & 8.116 \\ \hline
    Polish & 0.090 & 0.080 & 8.318 & 8.751$^{***}$ & 8.670$^{***}$ & 8.399 \\ \hline
    Portuguese & 0.186 & 0.155 & 7.907 & 8.921$^{***}$ & 8.642$^{***}$ & 8.187 \\ \hline
    Romanian & 0.175 & 0.145 & 8.682 & 8.658 & 8.934$^{***}$ & 8.406 \\ \hline
    Slovenian & 0.093 & 0.113 & 8.046 & 7.983 & 8.177$^{***}$ & 7.853 \\ \hline
    Spanish & 0.235 & 0.214 & 7.898 & 8.961$^{***}$ & 8.691$^{***}$ & 8.168 \\ \hline
    Swedish & 0.088 & 0.082 & 8.262$^{*}$ & 8.147 & 8.328$^{***}$ & 8.081 \\ \hline
    {\bf Overall} & {\bf 1 of 3} & {\bf 2 of 3} & {\bf 3 of 17} & {\bf 14 of 17} & {\bf 20 of 20} & {\bf 0 of 20} \\ \hline

    \end{tabular}
    \caption{Semantic metrics for 25 languages, computed using mBERT and 10M tokens of Wikipedia text for each language. Asterisks denote significance at $^{*}p < 0.05$, $^{**}p < 0.01$, $^{***}p<0.001$. For the ``Overall'' row, we count the languages with a significant tendency towards one direction, out of the number of languages with statistical significance towards either direction (with $p < 0.05$ treated as significant).}
    \label{tab:mbert-semantic-stats}
\end{table*}

\subsection{Three contextual metrics}

We define three metrics based on contextual embeddings to measure various semantic aspects of word class flexibility. We start by generating contextual embeddings for each occurrence of every flexible lemma. For each lemma $l$, let $E_{n, l}$ and $E_{v, l}$ be the set of contextual embeddings for noun and verb instances of $l$. We define the {\em prototype noun vector} $\vec{p}_{n, l}$ of a lemma $l$ as the mean of embeddings across noun instances, and the {\em noun variation} $V_{n, l}$ as the mean Euclidean distance from each noun instance to the noun vector:
\begin{align}
    & \vec{p}_{n,l} = \frac{1}{|E_{n, l}|} \sum_{\vec{x} \in E_{n,l}} \vec{x} \\
    & V_{n, l} = \frac{1}{|E_{n, l}|} \sum_{\vec{x} \in E_{n, l}} ||\vec{x} - \vec{p}_{n, l}||
\end{align}

The {\em prototype verb vector} $\vec{p}_{v, l}$ and {\em verb variation} $V_{v, l}$ for a lemma $l$ are defined similarly:
\begin{align}
    & \vec{p}_{v,l} = \frac{1}{|E_{v, l}|} \sum_{\vec{x} \in E_{v,l}} \vec{x} \\
    & V_{v, l} = \frac{1}{|E_{v, l}|} \sum_{\vec{x} \in E_{v, l}} ||\vec{x} - \vec{p}_{v, l}||
\end{align}

Lemmas are included if they appear at least 30 times as nouns and 30 times as verbs. To avoid biasing the variation metric towards the majority class, we downsample the majority class to be of equal size as the minority class before computing the variation. The method does not filter out pairs of lemmas that are arguably homonyms rather than flexible (section \ref{sec:polysemy-homonymy}); we choose to include all of these instances rather than set an arbitrary cutoff threshold.

We now define language-level metrics to measure the asymmetries hypothesized in sections \ref{sec:related-work-directionality} and \ref{sec:related-work-semantic-shift}. The {\em noun-to-verb shift (NVS)} is the average cosine distance between the prototype noun and verb vectors for noun dominant lemmas, and the {\em verb-to-noun shift (VNS)} likewise for verb dominant lemmas:
\begin{align}
    & NVS = 1 - \mathbb{E}_{l \textrm{ noun-dominant}} [\cos(\vec{p}_{n, l}, \vec{p}_{v, l})] \\
    & VNS = 1 - \mathbb{E}_{l \textrm{ verb-dominant}} [\cos(\vec{p}_{n, l}, \vec{p}_{v, l})]
\end{align}

We define the {\em noun (verb) variation} of a language as the average of noun (verb) variations across all lemmas. Finally, define the {\em majority variation} of a language as the average of the variation of the dominant POS class, and the {\em minority variation} as the average variation of the smaller POS class, across all lemmas.

\begin{table*}
    \centering
    \begin{tabular}{|c|l||l|l||l|l||l|l|}
    \hline
    \textbf{Dataset} & \textbf{Model} & \textbf{N$\to$V shift} & \textbf{V$\to$N shift} & \textbf{\begin{tabular}[c]{@{}l@{}}Noun\\ variation\end{tabular}} & \textbf{\begin{tabular}[c]{@{}l@{}}Verb\\ variation\end{tabular}} & \textbf{\begin{tabular}[c]{@{}l@{}}Majority\\ variation\end{tabular}} & \textbf{\begin{tabular}[c]{@{}l@{}}Minority\\ variation\end{tabular}} \\ \hline
    \multirow{4}{*}{BNC} & ELMo & 0.389$^{*}$  & 0.357 & 20.261  & 20.455 & 20.329 & 20.388  \\ \cline{2-8}
    & BERT & 0.122$^{*}$ & 0.112 & 9.015 & 9.074 & 9.100$^{***}$ & 8.989 \\ \cline{2-8}
    & mBERT & 0.189$^{*}$ & 0.169 & 7.211 & 8.401$^{***}$ & 7.875$^{**}$ & 7.717 \\ \cline{2-8}
    & XLM-R & 0.004 & 0.005 & 2.058 & 2.374$^{***}$ & 2.262 & 2.170 \\ \hline \hline
    \multirow{4}{*}{Wikipedia}& ELMo & 0.339$^{***}$  & 0.330 & 22.556 & 22.521 & 22.463 & 22.614$^{*}$  \\ \cline{2-8}
    & BERT & 0.120$^{***}$ & 0.100 & 9.218$^{***}$ & 8.944 & 9.118$^{**}$ & 9.044 \\ \cline{2-8}
    & mBERT & 0.175$^{*}$ & 0.160 & 8.035 & 8.624$^{***}$ & 8.390$^{***}$ & 8.268 \\ \cline{2-8}
    & XLM-R & 0.004$^{**}$ & 0.003 & 1.966 & 1.954 & 1.946 & 1.974 \\ \hline
    \end{tabular}
    \caption{Comparison of semantic models on BNC and Wikipedia datasets (English), computed using several different language models. Asterisks denote significance at $^{*}p < 0.05$, $^{**}p < 0.01$, $^{***}p<0.001$.}
    \label{tab:english-sem-compare}
\end{table*}

\section{Results}

\subsection{Identifying flexible lemmas}

Of the 37 languages in UD with at least 100k tokens; in 27 of them, at least 2.5\% of verb and noun lemmas are flexible, which we take to indicate that word class flexibility exists in the language (Table \ref{tab:ud-frequency}). The lemma merging algorithm is crucial for identifying word class flexibility: only 6 of the 37 languages pass the 2.5\% flexibility threshold using the default lemma annotations provided in UD\footnote{Chinese, Danish, English, Hebrew, Indonesian, and Japanese pass the flexibility threshold without the lemma merging algorithm.}. Languages differ in their prevalence of word class flexibility, but every language in our sample has higher verb flexibility than noun flexibility.

\subsection{Asymmetry in semantic metrics}

Table \ref{tab:mbert-semantic-stats} shows the values of the three metrics, computed using mBERT and Wikipedia data for 25 languages\footnote{We exclude 2 of the 27 languages that we identify word class flexibility. Old Russian was excluded because it is not supported by mBERT; Korean is excluded because the lemma annotations deviate from the standard UD format.}. For testing significance, we use the unpaired Student's t-test to compare N-V versus V-N shift, and the paired Student's t-test for the other two metrics\footnote{We do not apply the Bonferroni correction for multiple comparisons, because we  make claims for trends across all languages, and not for any specific languages.}. The key findings are as follows:

\begin{enumerate}
    \item {\bf Asymmetry in semantic shift.} In English, N-V shift is greater than V-N shift, in agreement with \citet{Bauer2005}. However, this pattern does not hold in general: there is no significant difference in either direction in most languages, and two languages exhibit a difference in the opposite direction as English.
    \item {\bf Asymmetry in semantic variation between noun and verb usages.} Of the 17 languages with a statistically significant difference in noun versus verb variation, 14 of them have greater verb variation than noun variation.
    \item {\bf Asymmetry in semantic variation between majority and minority classes.} All of the 20 languages with a statistically significant difference in majority and minority variation have greater majority variation.
\end{enumerate}

\subsection{Model robustness}

Next, we assess the robustness of our metrics with respect to choices of corpus and language model. Robustness is desirable because it gives confidence that our models capture true linguistic tendencies, rather than artifacts of our datasets or the models themselves. We compute the three semantic metrics on the BNC and Wikipedia datasets, using all 4 contextual language models: ELMo, BERT, mBERT, and XLM-R. Table \ref{tab:english-sem-compare} summarizes the results from this experiment.

We find that in almost every case where there is a statistically significant difference, all models agree on the direction of the difference. One exception is that noun variation is greater when computed using Wikipedia data than when using the BNC corpus. Wikipedia has many instances of nouns used in technical senses (e.g., \textit{ring} is a technical term in mathematics and chemistry), whereas similar nonfiction text is less common in the BNC corpus.

\section{Discussion}

\subsection{Frequency asymmetry}

Every language in our sample has verb flexibility greater than noun flexibility. The reasons for this asymmetry are unclear, but may be due to semantic differences between nouns and verbs. We note that every language in our sample has more noun lemmas than verb lemmas, a pattern that was also attested by \citet{more-nouns-than-verbs}, although this does not provide an explanation of the observed phenomenon. We leave further exploration of the flexibility asymmetry to future work.

\subsection{Implications for theories of flexibility}

There is a strong cross-linguistic tendency for the majority word class of a flexible lemma to exhibit more semantic variation than the minority class. In other words, the frequency and semantic variation criteria of determining the base of a conversion pair agree more than at chance. This supports the analysis of word class flexibility as a directional process of conversion, as opposed to underspecification (section \ref{sec:related-work-directionality})\footnote{Since 18 of the 25 languages for which semantic metrics were calculated are Indo-European, it is unclear whether these results generalize to non-Indo-European languages.}. Within a flexible lemma, verbs exhibit more semantic variation than nouns. It is attested across many languages that nouns are more physically salient, while verbs have more complex event and argument structure, and are harder for children to acquire than nouns \citep{gentner-1982, imai-2008}. Thus, verbs are expected to have greater semantic variation than nouns, which our results confirm. More importantly, for our purposes, this metric serves as a control for the previous metric. Flexible lemmas are more likely to be noun-dominant than verb-dominant, so could the majority and minority variation simply be proxies for noun and verb variation, respectively? In fact, we observe greater verb than noun variation, so this cannot be the case.

Finally, as suggested by \citet{Bauer2005}, we find evidence in English that N-V flexibility involves more semantic shift than V-N flexibility, and the pattern is consistent across multiple models and datasets (Table \ref{tab:english-sem-compare}). However, this pattern is idiosyncratic to English and not a cross-linguistic tendency. It is thus instructive to analyze multiple languages in studying word class flexibility, as one can easily be misled by English-based analyses.

\section{Conclusion}

We use contextual language models to examine shared tendencies in word class flexibility across languages. We find that the majority class often exhibits more semantic variation than the minority class, supporting the view that word class flexibility is a directional process. We also find that in English, noun-to-verb flexibility is associated with more semantic shift than verb-to-noun flexibility, but this is not the case for most languages.

Our probing task reveals that the upper layers of BERT contextual embeddings best reflect human judgment of semantic similarity. We obtain similar results in different datasets and language models in English that support the robustness of our method. This work demonstrates the utility of deep contextualized models in linguistic typology, especially for characterizing cross-linguistic semantic phenomena that are otherwise difficult to quantify.

\section*{Acknowledgments}

YX is funded through a Connaught New Researcher Award, a NSERC Discovery Grant RGPIN-2018-05872, and a SSHRC Insight Grant \#435190272.

\bibliography{emnlp-ijcnlp-2019}
\bibliographystyle{acl_natbib}

\clearpage
\appendix
%\section{Supplemental Material}

% \section{Proof of frequency asymmetry argument}
% \label{appendix:asymmetry-proof}

% {\bf Claim:} Any language with more noun than verb lemmas should have greater verb flexibility than noun flexibility.

% {\em Proof.}  Let us define $N$ as number of inflexible noun lemmas, $V$ as the number of inflexible verb lemmas, and $F$ as the number of flexible lemmas that occur as both nouns and verbs.

% Let us further subdivide $F = FN + FV$, where $FN$ is the number of flexible noun dominant lemmas, and $FV$ is the set of flexible verb dominant lemmas. We do not assume any prior knowledge about whether a flexible lemma is noun or verb dominant, so assume $FN = FV$.

% We defined noun flexibility as $\frac{FN}{FN + N}$ and verb flexibility as $\frac{FV}{FV + V}$. We have assumed that $N > V$ and $FN = FV$, therefore it follows that $\frac{FN}{FN+N} < \frac{FV}{FV+V}$. $\square$

\begin{table*}[h]
    \centering
\begin{tabular}{||l|l|l|l||l|l|l|l||l|l|l|l||}
\hline
\textbf{Word} & \textbf{$|N|$} & \textbf{$|V|$} & \textbf{Sim} & \textbf{Word} & \textbf{$|N|$} & \textbf{$|V|$} & \textbf{Sim} & \textbf{Word} & \textbf{$|N|$} & \textbf{$|V|$} & \textbf{Sim} \\ \hline
aim & 137 & 98 & 2.0 & change & 889 & 858 & 1.6 & force & 470 & 188 & 0.8 \\ \hline
answer & 480 & 335 & 2.0 & claim & 222 & 239 & 1.6 & grant & 108 & 87 & 0.8 \\ \hline
attempt & 302 & 214 & 2.0 & cut & 92 & 488 & 1.6 & note & 287 & 361 & 0.8 \\ \hline
care & 403 & 249 & 2.0 & demand & 169 & 142 & 1.6 & sense & 536 & 88 & 0.8 \\ \hline
control & 519 & 179 & 2.0 & design & 246 & 153 & 1.6 & tear & 124 & 89 & 0.8 \\ \hline
cost & 234 & 192 & 2.0 & experience & 522 & 150 & 1.6 & account & 337 & 122 & 0.6 \\ \hline
count & 143 & 220 & 2.0 & hope & 114 & 571 & 1.6 & act & 644 & 268 & 0.6 \\ \hline
damage & 270 & 82 & 2.0 & increase & 252 & 399 & 1.6 & back & 764 & 88 & 0.6 \\ \hline
dance & 81 & 97 & 2.0 & judge & 80 & 96 & 1.6 & face & 1185 & 281 & 0.6 \\ \hline
doubt & 261 & 132 & 2.0 & limit & 125 & 134 & 1.6 & hold & 130 & 1251 & 0.6 \\ \hline
drink & 456 & 315 & 2.0 & load & 230 & 87 & 1.6 & land & 393 & 123 & 0.6 \\ \hline
end & 1171 & 244 & 2.0 & offer & 93 & 489 & 1.6 & lift & 100 & 165 & 0.6 \\ \hline
escape & 95 & 111 & 2.0 & rise & 164 & 283 & 1.6 & matter & 572 & 294 & 0.6 \\ \hline
estimate & 96 & 118 & 2.0 & smoke & 128 & 100 & 1.6 & order & 841 & 133 & 0.6 \\ \hline
fear & 209 & 99 & 2.0 & start & 159 & 1269 & 1.6 & place & 1643 & 341 & 0.6 \\ \hline
glance & 101 & 161 & 2.0 & step & 401 & 167 & 1.6 & press & 130 & 188 & 0.6 \\ \hline
help & 200 & 897 & 2.0 & study & 1037 & 211 & 1.6 & roll & 135 & 201 & 0.6 \\ \hline
influence & 204 & 150 & 2.0 & support & 290 & 292 & 1.6 & sort & 1613 & 216 & 0.6 \\ \hline
lack & 194 & 107 & 2.0 & trust & 90 & 126 & 1.6 & fire & 444 & 89 & 0.4 \\ \hline
link & 147 & 176 & 2.0 & waste & 103 & 98 & 1.6 & form & 1272 & 354 & 0.4 \\ \hline
love & 495 & 573 & 2.0 & work & 1665 & 1593 & 1.6 & notice & 115 & 387 & 0.4 \\ \hline
move & 131 & 1272 & 2.0 & base & 109 & 378 & 1.4 & play & 185 & 1093 & 0.4 \\ \hline
name & 960 & 112 & 2.0 & cover & 137 & 399 & 1.4 & turn & 226 & 1566 & 0.4 \\ \hline
need & 587 & 2350 & 2.0 & plant & 591 & 82 & 1.4 & wave & 402 & 120 & 0.4 \\ \hline
phone & 382 & 238 & 2.0 & run & 152 & 999 & 1.4 & cross & 102 & 215 & 0.2 \\ \hline
plan & 321 & 161 & 2.0 & stress & 159 & 106 & 1.4 & deal & 191 & 315 & 0.2 \\ \hline
question & 1285 & 96 & 2.0 & approach & 409 & 175 & 1.2 & hand & 1765 & 127 & 0.2 \\ \hline
rain & 182 & 92 & 2.0 & cause & 237 & 530 & 1.2 & present & 219 & 353 & 0.2 \\ \hline
result & 752 & 206 & 2.0 & match & 110 & 123 & 1.2 & set & 387 & 652 & 0.2 \\ \hline
return & 138 & 441 & 2.0 & miss & 320 & 410 & 1.2 & share & 104 & 232 & 0.2 \\ \hline
search & 215 & 163 & 2.0 & process & 720 & 91 & 1.2 & sign & 284 & 121 & 0.2 \\ \hline
sleep & 171 & 291 & 2.0 & shift & 96 & 104 & 1.2 & suit & 162 & 108 & 0.2 \\ \hline
smell & 141 & 149 & 2.0 & show & 132 & 1843 & 1.2 & wind & 189 & 82 & 0.2 \\ \hline
smile & 211 & 422 & 2.0 & sound & 313 & 496 & 1.2 & address & 257 & 148 & 0.0 \\ \hline
talk & 119 & 1302 & 2.0 & dress & 191 & 196 & 1.0 & bear & 110 & 394 & 0.0 \\ \hline
use & 791 & 2801 & 2.0 & lead & 107 & 716 & 1.0 & head & 1355 & 96 & 0.0 \\ \hline
view & 811 & 102 & 2.0 & light & 669 & 124 & 1.0 & mind & 736 & 620 & 0.0 \\ \hline
visit & 136 & 203 & 2.0 & look & 699 & 5893 & 1.0 & park & 179 & 105 & 0.0 \\ \hline
vote & 124 & 93 & 2.0 & mark & 562 & 198 & 1.0 & point & 1534 & 469 & 0.0 \\ \hline
walk & 144 & 914 & 2.0 & measure & 226 & 223 & 1.0 & ring & 185 & 387 & 0.0 \\ \hline
dream & 254 & 107 & 1.8 & rest & 414 & 132 & 1.0 & square & 225 & 82 & 0.0 \\ \hline
record & 1057 & 276 & 1.8 & tie & 82 & 112 & 1.0 & state & 471 & 156 & 0.0 \\ \hline
report & 313 & 331 & 1.8 & break & 117 & 519 & 0.8 & stick & 109 & 294 & 0.0 \\ \hline
test & 273 & 126 & 1.8 & charge & 392 & 115 & 0.8 & store & 95 & 158 & 0.0 \\ \hline
touch & 145 & 271 & 1.8 & drive & 88 & 476 & 0.8 & train & 224 & 94 & 0.0 \\ \hline
call & 209 & 1558 & 1.6 & focus & 92 & 168 & 0.8 & watch & 119 & 940 & 0.0 \\ \hline
\end{tabular}
    \caption{138 flexible words in English (top in BNC corpus) and human similarity scores, average of 5 ratings.}
    \label{tab:english-lemma-ratings}
\end{table*}

\end{document}